\documentclass[11pt,onecolumn,a4paper]{article}

\usepackage{amsmath}
\usepackage{amsfonts}
\usepackage{amssymb}
\usepackage{amsthm}

\usepackage[ansinew]{inputenc}
\usepackage{graphicx,color}
\usepackage{url}
\usepackage{latexsym,enumerate}
\usepackage{booktabs}
\usepackage{framed}
\usepackage{subfigure}
\usepackage{multirow} 
\usepackage{color}
\usepackage{colortbl}
\definecolor{gray1}{rgb}{0.8,0.8,0.8}
\definecolor{gray2}{rgb}{0.95,0.95,0.95}

\begin{document}

\title{Skilled Impostor Attacks Against Fingerprint Verification Systems And Its Remedy}

\author{Carsten Gottschlich\thanks{Institute for Mathematical Stochastics,
University of G\"ottingen,
Goldschmidtstr. 7, 37077 G\"ottingen, Germany.
Email: gottschlich@math.uni-goettingen.de}}

\date{}

\maketitle


\begin{abstract}

Fingerprint verification systems are becoming ubiquitous in everyday life. 
This trend is propelled especially by the proliferation
of mobile devices with fingerprint sensors such as smartphones and tablet computers,
and fingerprint verification is increasingly applied for 
authenticating financial transactions.
In this study we describe a novel attack vector against fingerprint verification systems 
which we coin skilled impostor attack.
We show that existing protocols for performance evaluation of
fingerprint verification systems are flawed 
and as a consequence of this, the system's real vulnerability is systematically underestimated.
We examine a scenario in which a fingerprint verification system
is tuned to operate at false acceptance rate of $0.1\%$
using the traditional verification protocols with random impostors (zero-effort attacks).
We demonstrate that an active and intelligent attacker 
can achieve a chance of success in the area of $89\%$ or more against this system 
by performing skilled impostor attacks.
We describe a new protocol for evaluating fingerprint verification performance
in order to improve the assessment of potential and limitations
of fingerprint recognition systems.
This new evaluation protocol enables a more informed decision 
concerning the operating threshold in practical applications
and the respective trade-off between security (low false acceptance rates) 
and usability (low false rejection rates).
The skilled impostor attack is a general attack concept 
which is independent of specific databases or comparison algorithms.
The proposed protocol relying on skilled impostor attacks can directly be applied
for evaluating the verification performance of other biometric modalities 
such as e.g. iris, face, ear, finger vein, gait or speaker recognition.

\end{abstract}

\section*{Keywords}

Biometric performance evaluation, fingerprint verification, 
fingerprint recognition, evaluation protocol, 
spoof attacks, zero-effort attacks, skilled impostor attacks, 
multi-factor authentication.

\section{Introduction and Motivation}

The use of fingerprints for establishing the identity of a person has a long history. 
Impressions discovered on earthenware in northwest China are estimated to be 6000 years old 
(see Chapter 1 in \cite{FingerprintSourcebook2011}).
In 1892, Galton \cite{Galton1892} published an influential book in which he explored 
the uniqueness and persistence of fingerprints.
Law enforcement agencies have been employing fingermarks left at crime scenes for identifying suspects
routinely and successfully in forensic investigations
for more than a century \cite{FingerprintsAndOtherRidgeSkinImpressions2004}.

Beyond forensic and governmental applications like border control,
fingerprints are increasingly used in commercial applications 
for check-in at workplaces or libraries,
for access control at amusement parks or zoos,
and for unlocking mobile computers like laptops, tablet computers or smartphones.
The trend of fingerprint sensors on mobile devices drives the pervasion
of everyday life with fingerprint technology.
Mobile phones with fingerprint sensors have been developed years ago 
by Siemens (1998),
by Sagem (2000),
by Fujitsu (2003),
by Motorola (2008) and by many other companies,
however, especially Apple and Samsung propel the widespread use of fingerprints on smartphones.
The European Association for Biometrics considers the release of the iPhone 5s in September 2013
to herald a paradigm shift \cite{EABOniPhone5s}.
317 million units of fingerprint-enabled devices are expected to be shipped by the end of 2014
and IHS predicts this number to increase to 1.4 billion units in 2020 \cite{IHSPrediction2014}.

A second emerging trend is the authentication of payments and other financial transactions by biometrics.
In addition to mobile payment services by Apple, Samsung and others, a credit card with a fingerprint scanner 
has been developed by Zwipe and MasterCard \cite{ZwipeMastercard2014}.

Traditional two-factor authentication relies on 
a possession factor such as a card for an automated teller machines (ATM) or a credit card and
on a knowledge factor such as a personal identification number (PIN) or a password.
A biometric factor such as a fingerprint or a signature 
can be used as a third factor to increase the security of a system.
For example, in an ATM scenario, a customer could be asked for an ATM card, a PIN and a fingerprint (three-factor authentication)
before being able to withdraw money.
The aforementioned credit card with a fingerprint sensor enables 
two-factor authentication (possession factor and biometric factor).
In this specific application, fingerprint recognition replaces a PIN \cite{BonneauHerleyVanoorschotStajano2012}.
Choices for certain types of multi-factor authentication 
involve always a trade-off between security and convenience.

The prospect of ubiquity of devices with fingerprint sensors
paired with the trend towards authenticating financial transactions by fingerprint recognition
creates a strong motivation to analyze the security and vulnerability of fingerprint recognition systems
as well as associated privacy concerns.

\section{Attacks} 

As fingerprinting is becoming ubiquitous and together with the rise of commercial applications relying on fingerprints,
it is also becoming increasingly attractive for criminals to attack fingerprint recognition systems.

In this section, we consider three different kind of attacks on fingerprint verification systems:
spoof attacks \cite{GottschlichMarascoYangCukic2014,AkhtarMichelonForesti2014,SousedikBusch2014,MarascoRoss2014}, 
zero-effort attacks and skilled impostor attacks.
The first two types of attacks are well known, 
the third type has not been considered by the research community so far.

All three types of attacks are so-called direct attacks \cite{MarascoRoss2014}
which means that point of attack is the fingerprint sensor.
Typically, the sensor is assumed to be publicly accessible.

Further kinds of attacks  
are conceivable against internal modules and communication channels of a fingerprint system, 
e.g. an attacker could try to modify a template stored in a database or attempt to replace 
the feature extraction module by a malicious software. 
This category of attacks is called indirect attacks and not considered here.

\subsection{Spoof Attacks}

A spoof finger is a fake finger made from artificial material such as e.g. latex, wood glue, silicone, or gelatin.
The spoof finger is presented to the sensor in order to impersonate a genuine user of the system,
therefore spoof attacks belong to the category of presentation attacks \cite{SousedikBusch2014}. 
We can distinguish between three different ways to fabricate fake fingers.
Spoof fingers can be produced with the cooperation of the genuine user 
as a so-called direct cast, without cooperation (indirect cast) 
or from a synthetic fingerprint image.

In the cooperative scenario, the person presses his or her finger into a mold 
made of e.g. modeling putty like plasticine or candle wax.
The spoof is created by inserting e.g. silicone or gelatin into the mold.
An example for a cooperative attack on a fingerprint system was reported
by the police of Ferraz de Vasconcelos, near Sao Paulo, Brazil, in March 2013.
A doctor was caught on camera forging the check-in of absent co-workers.
She had six silicone fingers with her at the time of her arrest.
Subsequent investigations revealed that as many as 300 civil servants
in that town were receiving pay packets without going to work \cite{DoctorSpoofBBC}.

In the non-cooperative scenario, a fingerprint from the target of impersonation
is acquired without agreement or help of that person,
e.g. by lifting a latent fingerprint from a glass previously touched by that person.
Typical steps of fingerprint image preprocessing \cite{Gottschlich2010PhD}
can be applied to improve the visualization of the latent fingerprint:
Segmentation into foreground and background \cite{ThaiHuckemannGottschlich2015}, 
orientation field estimation
(e.g. by the line sensor method \cite{GottschlichMihailescuMunk2009}) and
image enhancement \cite{Gottschlich2012,GottschlichSchoenlieb2012}.
The enhanced image can be printed to obtain a mold \cite{SousedikBusch2014}.
This kind of attack has been demonstrated by members 
of the Chaos Computer Club in Germany against the Apple iPhone 5S
immediately after its release in September 2013 \cite{CCC}.
The same approach was also successful to show the vulnerability of the Samsung Galaxy S5
to spoof fingers.
An additional attack vector in the non-cooperative scenario 
is the reconstruction of a fingerprint image from a minutiae template
which can be achieved e.g. using amplitude- and frequency-modulated (AM-FM) functions \cite{LarkinFletcher2007}.
A list of references to further reconstruction methods is provided by \cite{GottschlichHuckemann2014}.
Consequently, if an attacker can steal a database of minutiae templates from a server or cloud storage, 
it is possible to reconstruct the corresponding fingerprint images and create spoof fingers from those.

A third possibility is to generate a synthetic fingerprint image and create a spoof finger from it.
It has recently been shown that existing methods for generating artificial fingerprint images 
tend to produce synthetic fingerprints with unrealistic minutiae configurations \cite{GottschlichHuckemann2014}.
Analyzing and representing templates by minutiae histograms (MHs) \cite{GottschlichHuckemann2014}
enables to separate between real and synthetic fingerprints.
More specifically, a template can be classified as real or synthetic by computing 
the earth mover's distance \cite{GottschlichSchuhmacher2014}
between the template's MH and the mean MHs of real and synthetic fingerprints.
Despite being unrealistic, an attacker could attempt to enroll a spoof finger of a synthetic fingerprint
to fingerprint recognition system and if successful, this virtual identity could also be shared by several attackers.

\subsection{Zero-Effort Attacks}

A zero-effort attack against a fingerprint verification system
is to present a real and alive finger to the sensor
in combination with the claim of an identity not associated with this finger.
For example, let us assume that Alice is a legitimate user 
of a fingerprint verification system and costumer of a bank 
using three-factor authentication (bank card, PIN and fingerprint)

During the enrollment phase,
she was registered to the database of the bank and e.g. a minutiae template
extracted from a fingerprint of Alice 
was stored together with information about her identity.

In order to draw money from an ATM,
Alice presents her bank card (which makes an implicit claim regarding her identity),
she types in her PIN and presents her finger to the sensor. 
Alice is a legitimate user of the system and verification attempt
is also called ``genuine recognition attempt''.

Let us assume that an attacker called Mallory somehow attains possession
of Alice's bank card and learns her PIN.
If Mallory now uses one of her own fingers together with the bank card and PIN from Alice,
trying to draw money and debit Alice's bank account, 
then Mallory performs a zero-effort attack on the fingerprint recognition system.
Traditionally, this attack is known as 
``impostor recognition attempt'' in the literature \cite{HandbookFingerprintRecognition2009}.

\subsection{Skilled Impostor Attacks}

\begin{figure}
\begin{center}
\includegraphics[width=0.7\textwidth]{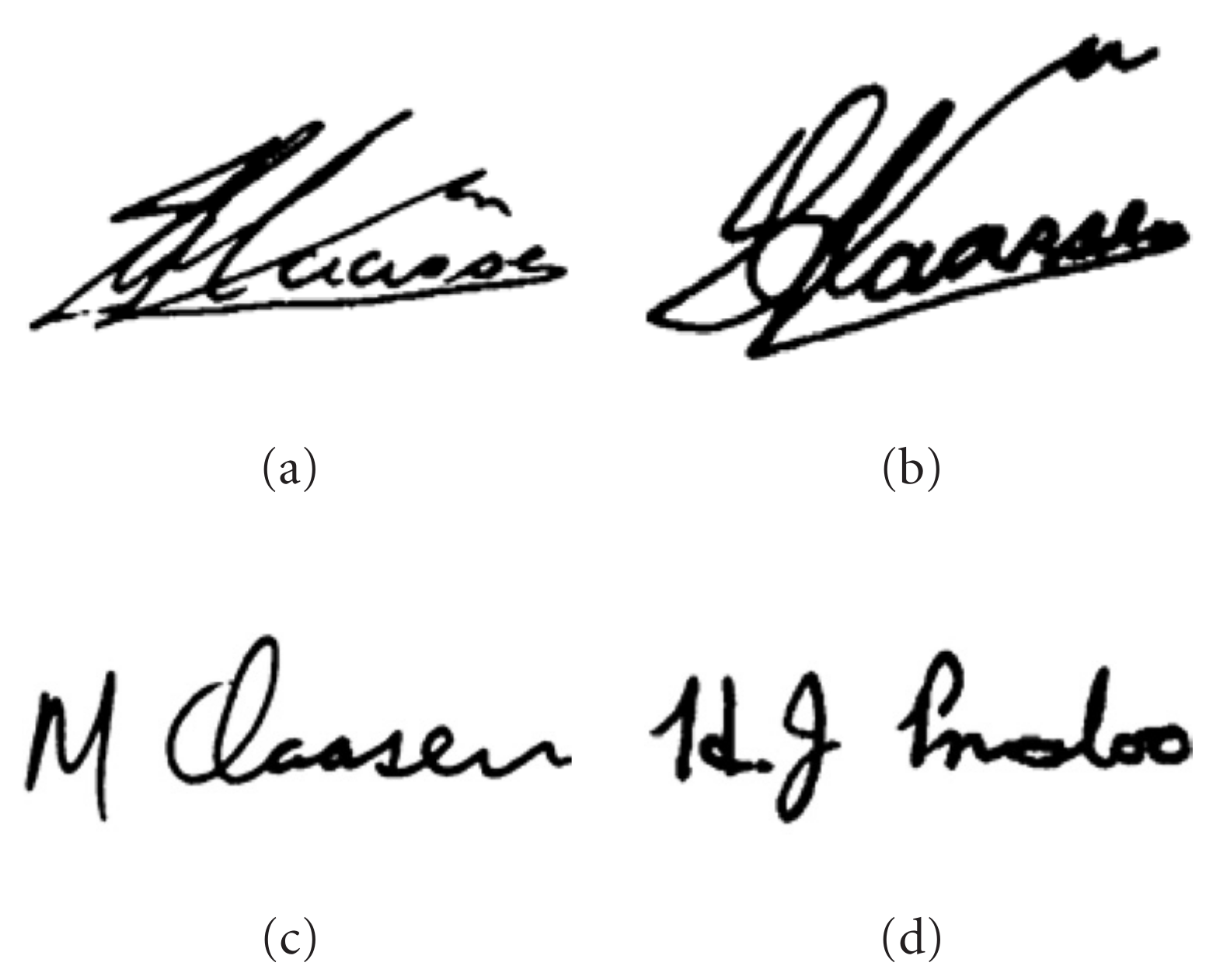}
\end{center}
\caption{The signature of a genuine user is depicted in (a).
         A skilled forgery is shown in (b) and a casual forgery in (c). 
				 An example of a random forgery is illustrated in (d).
				 This Figure is reproduced from the work of Coetzer \textit{et al.} \cite{CoetzerHerbstDupreez2004}.
         See especially Figures 1-3 in \cite{CoetzerHerbstDupreez2004}.
}
\label{fig:signature}
\end{figure}

A comparison algorithm computes a similarity (score) between
two fingerprint feature sets, e.g. between Alice's minutiae template 
in the database and a second minutiae template extracted from the finger currently placed on the sensor.
Based on the score $s$ and the threshold $t$, 
the system decides to accept, if $s \geq t$,
or to reject otherwise.

In the described verification scenario (one-to-one comparison) at the ATM, 
a fingerprint recognition system can make two types of errors. 
It can produce a false accept (e.g. accept Mallory's finger with the claim to be Alice's) 
or a false reject (e.g. reject Alice although she is a legitimate user 
and she placed the same finger on the sensor that has been enrolled earlier).

\begin{figure}
\begin{center}
\includegraphics[width=0.7\textwidth]{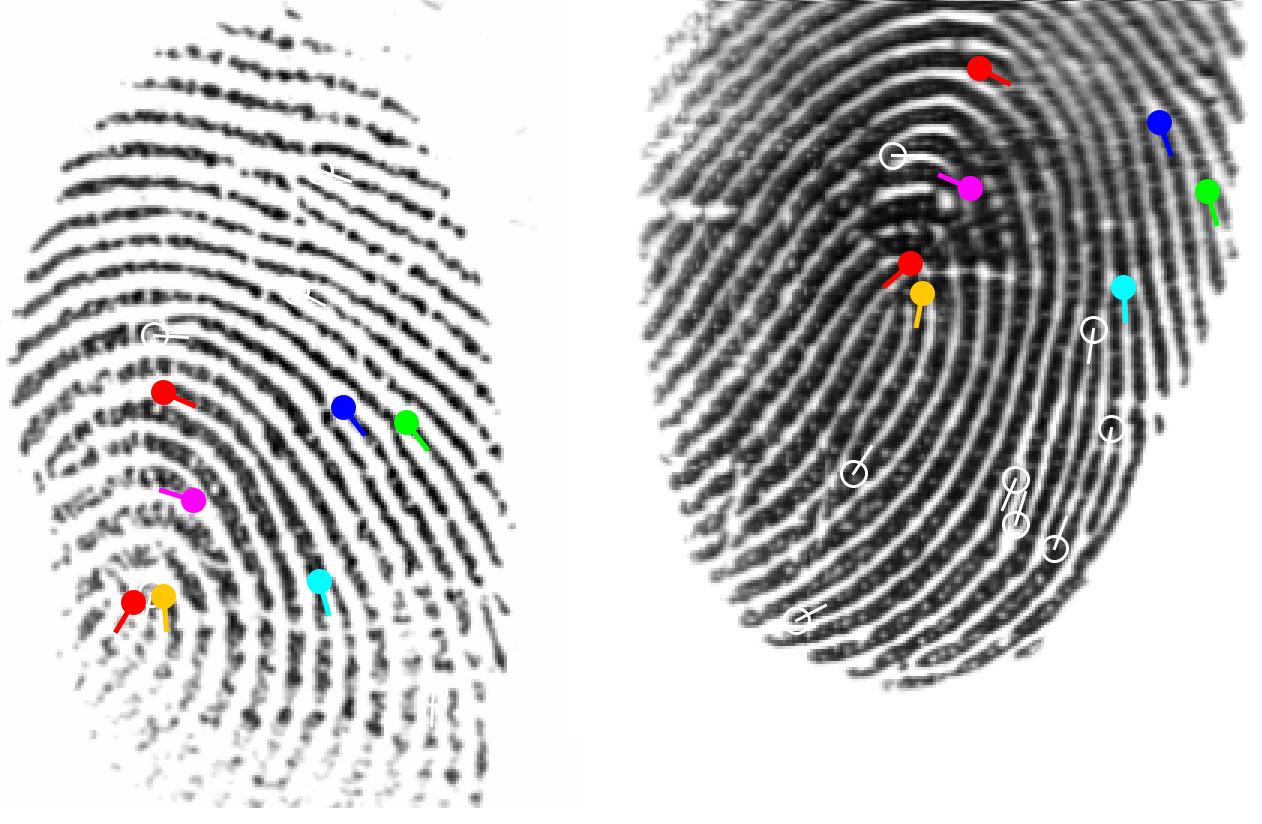}
\end{center}
\caption{This skilled impostor recognition attempt 
shows finger no. 28 impression 7 from FVC2002 DB1 (left)
which is attacked by finger no. 20 impression 1 from FVC2004 DB1.
The automatic comparison algorithm paired 7 minutiae (highlighted in color, unpaired minutiae are shown in white)
and computed a similarity score of 120.
For a human observer it is clearly discernible that the two fingerprints 
originate from two different fingers.
For example, counting the intersecting ridge lines between the 
orange and cyan minutiae results in four ridges for the finger on the left 
and seven for the finger on the right.
Additionally, for the print on the right, there is a minutia clearly visible 
on the left beneath the cyan minutia which does not exist in the print on the left.
}
\label{fig:skilledImpostor}
\end{figure}

The operational threshold $t$ of fingerprint verification system
can be chosen to balance the two goals security (few false accepts)
and usability (few false rejects) for an application scenario.
Let us assume the ATM system has a false accept rate of $0.1\%$ 
measured against the traditional zero-effort attacks.
Let us further assume that the ATM system implements a perfect liveness detection method
which defends the system against all spoof attacks with $100\%$ accuracy.
How can Mallory attack such a system, if she knows Alice's fingerprint,
but a spoof attack is not possible?
If Mallory is part of a criminal organization, 
she can proceed in the following way to optimize the chances of a successful attack.
She acquires the fingerprints of all accomplices 
and she uses a fingerprint comparison algorithm to find 
the fingerprint which is most similar to Alice's fingerprint.
Say, this is one of Sybil's fingers.
Then, Sybil attacks the ATM using her selected finger together with Alice's bank card and PIN.
We denote this kind of attack as skilled impostor attack 
in allusion to skilled forgeries in signature verification (see \cite{CoetzerHerbstDupreez2004} 
and Figure \ref{fig:signature}). 
An example of a skilled impostor attack for fingerprint recognition is shown in Figure \ref{fig:skilledImpostor}.
You can imagine that fingerprint in Figure \ref{fig:skilledImpostor} on the left belongs to Alice 
and print in Figure \ref{fig:skilledImpostor} on the right belongs to the skilled impostor Sybil.
Scrutinizing the details in Figure \ref{fig:skilledImpostor} it is discernible 
that the figure depicts two fingerprints originating from two different fingers.
At the same time, it is very understandable why an algorithm 
can compute a high similarity value on comparing these two fingerprints
which ultimately can result in a false accept.

The assumption that Mallory is not acting on her own, but with accomplices is quite realistic
as many news reports confirm.
E.g. in October 2011 a break-in occurred at grocery store in Goettingen, Germany \cite{GoettingerTageblattSkimming2011}.
Only later the police learned that the main objective of the break-in was not to steal goods,
but to manipulate the bank card reader at the cash point.
The forged readers recorded bank card data and PINs between the 10th of October
and the 4th of November.
The criminals took possession of the data by a second break-in 
in the night from the 4th to 5th November.
One and a half day later, the first money withdrawals took place in Mexico and the USA.
This is an example of a so-called skimming case.
Identity theft and skimming are two very relevant types of credit card fraud
and a possible countermeasure is three-factor authentication 
with fingerprint verification.

In the next section, 
we describe the traditional protocol and we propose a new protocol for evaluating the performance of fingerprint verification systems
and afterwards  
we examine the chances of successful skilled impostor attacks
using publicly available fingerprint databases and state-of-the-art software for fingerprint verification.

\section{Protocols} 

Let us assume that we have a fingerprint database containing images of $n$ different fingers
denoted as $f_1, ..., f_n$ 
and we have $m$ different impressions for each finger which are given in a certain order.
For each fingerprint $f_i$ and its $j$-th impression, we denote by $v_{ij}$ 
the corresponding feature vector which can be a fingerprint image, a minutiae template or some other feature.
Further, we have a comparison algorithm $M$ (colloquially called a ``matcher'')
which takes feature vectors $v_{ij}$ and $v_{xy}$ of two fingerprints $f_i$ and $f_x$ as input
and computes a similarity value $s$ (``score'') as output.
A higher score corresponds to a greater similarity.
We define two protocols in the following way.
\\
\\
\textbf{Verification test with random impostors}

\textit{Genuine recognition attempts}:
For each of the $n$ fingers, 
each impression of a finger $f_i$ is compared to every other impression of the same finger $f_i$
while avoiding mirrored comparisons.
By mirrored comparison we denote in general that $v_{xy}$ is not compared to $v_{ij}$,
if previously $v_{ij}$ has been compared to $v_{xy}$. (For genuine recognition attempts, here $i = x$).
Hence, the first impression of finger $f_i$ is compared to all other impressions of finger $f_i$.
The second impression of finger $f_i$ is compared to all other impressions of finger $f_i$ except for the first.
The third impression of finger $f_i$ is compared to all other impressions of finger $f_i$ 
except for the first and second, and so on.
In total, this protocol carries out $c_{gen} = \frac{n \cdot m \cdot (m-1)}{2}$ genuine recognition attempts.

\textit{Impostor recognition attempts}:
For each of the $n$ fingers, the first $a$ impressions of finger $f_i$ 
are compared to first $a$ impressions 
of every other finger in the database
while avoiding mirrored comparisons.
This leads to $c_{imp} = \frac{n \cdot (n-1) \cdot a^2}{2}$ impostor recognition attempts.
\\
\\
\textbf{Verification test with skilled impostors}

\textit{Genuine recognition attempts}:
We perform exactly the same above described
$c_{gen} = \frac{n \cdot m \cdot (m-1)}{2}$ genuine recognition attempts.

\textit{Impostor recognition attempts}:
For each of the $n$ fingers and for each of the first $a$ impressions of finger $f_i$,
we search the $k$ most similar impressions considering 
the first $u$ impressions of
every other finger in the database. 
Similarity is measured by the comparison algorithm $M$.
Only the $k$ most similar impressions are taken into account as impostor recognition attempts.
In total, we have $c_{imp} = n \cdot a \cdot k$ skilled impostor recognition attempts.

As standard values for verification tests with skilled impostors 
we suggest $a = u = m$ which means that every image (or template) in the database
is considered as a target for impersonation and $k = 1$
which corresponds to an attack by the most skillful attacker.
Moreover, we suggest to include further available fingerprint databases for skilled impostor attacks
if a database contains only a small number of unique fingers $n$.
Additionally, for both protocols it should be checked whether $M$ is symmetric, i.e. $M(v_{ij}, v_{xy}) = s = M(v_{xy}, v_{ij})$
for all fingers and impressions. If $M$ is not symmetric, the otherwise avoided mirrored comparisons should also be carried out.

\section{Results}

The protocol for a verification test with random impostors
is - to the best knowledge of the author -
the only one which used in existing evaluations of fingerprint verification performance.
Prominent examples are the FVC protocol described in \cite{FVC2000} and the evaluations on the
GUC100 database \cite{GUC100}.

The experiments in this study are based on the publicly available fingerprint databases of 
FVC2000 \cite{FVC2000}, 
FVC2002 \cite{FVC2002},
FVC2004 \cite{FVC2004} and we use a commercial software as comparison algorithm.
First, we conduct a traditional verification test with random impostors.
We use FVC2002 DB1 as the bank database in our scenario, 
because it is a database with mostly good quality images.
The databases of FVC2004 contain many low-quality images 
and are employed for comparing fingerprint image enhancement algorithms (see e.g. 
\cite{Gottschlich2012,GottschlichSchoenlieb2012,BartunekNilssonSallbergClaesson2013,KocevarKotnikChowdhuryKacic2014})
using the traditional protocol of verification tests with random impostors.

Each FVC database comprises a training set of 10 fingers and a test set of $n = 100$ fingers.
For each finger, $m = 8$ impressions are available.
This results in $c_{gen} = \frac{n \cdot m \cdot (m-1)}{2} = 2800$ genuine recognition attempts.
The FVC protocol takes $t = 1$ impressions for random impostors into account,
therefore it carries out $c_{imp} = \frac{n \cdot (n-1) \cdot a^2}{2} = 4950$ impostor recognition attempts.

\begin{figure}
\begin{center}
\includegraphics[width=0.95\textwidth]{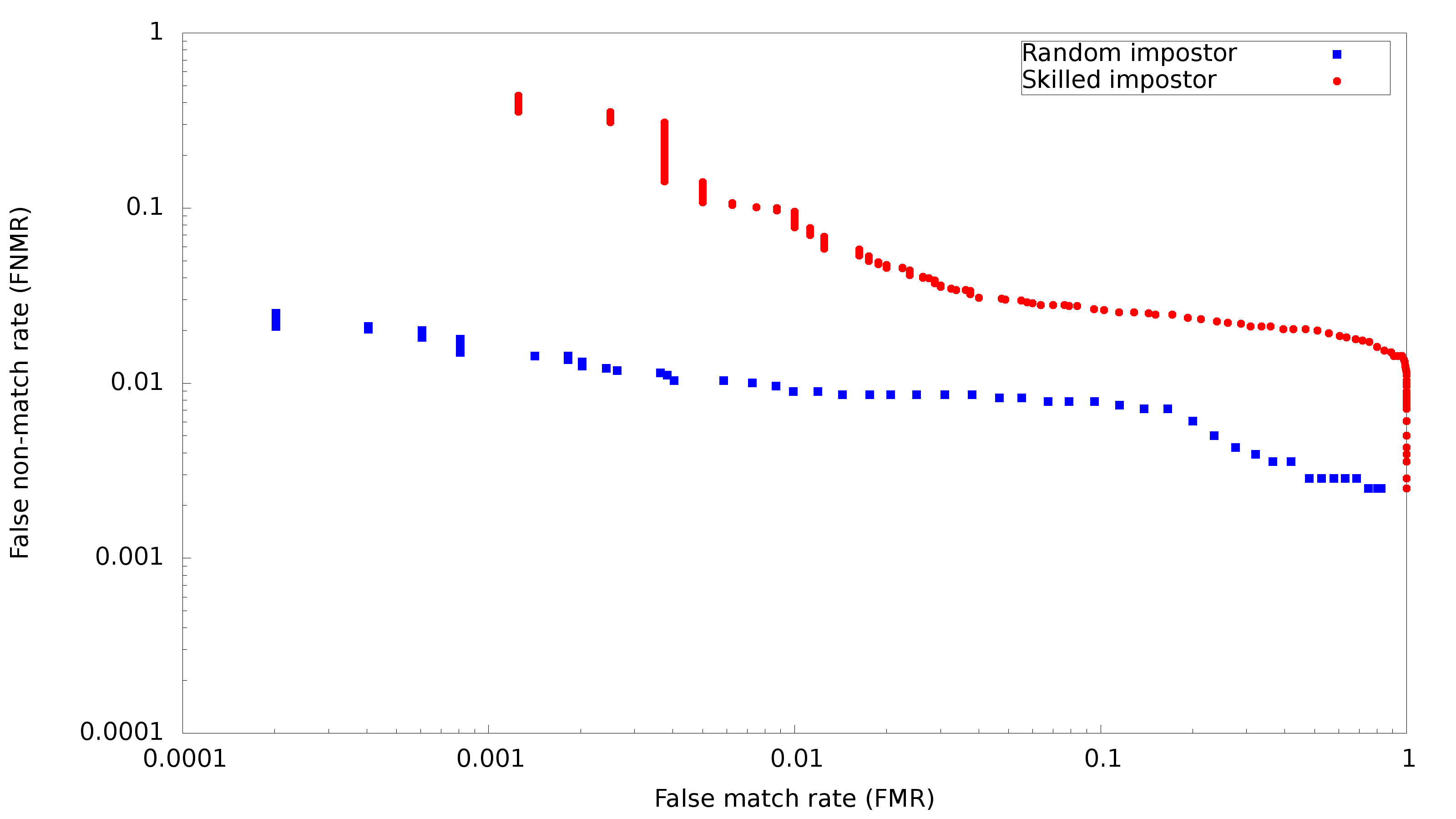}
\end{center}
\caption{Verification performance on FVC2002 DB1 using the same comparison algorithm 
and the same 2800 genuine recognition attempts.
The only difference between both experiments are the considered impostor recognition attempts:
4950 random impostors (blue) according to the FVC protocol and 800 skilled impostors (red).
}
\label{fig:det}
\end{figure}

The observed verification performance is visualized as a blue detection error tradeoff curve
in Figure \ref{fig:det}.
An equal error rate of $0.92\%$ is obtained at a threshold of $t = 33$.
The FMR1000 \cite{HandbookFingerprintRecognition2009} is observed at a threshold of $t = 48$,
i.e. the empirical probability for a false match is less or equal to $0.1\%$ at this threshold.

Let us return to the imagined ATM scenario with Alice's fingerprint in the bank database
and we assume that the bank implements this system for fingerprint verification 
with an operational threshold of $t = 48$, i.e. accepting all recognition attempts 
yielding a score of $s \geq 48$, and rejecting all other recognition attempts.
For the sake of argument, we neglect issues pertaining to 
repeatability and reproducibility of measurements in 
biometric performance testing \cite{WaymanPossoloMansfield2013},
and to put it simply, we say that Mallory's chances of a successful attack 
against the ATM fingerprint verification system 
using a random finger (e.g. one of her own ten fingers)
are one in a thousand.

Next, we perform a verification test with skilled impostors
and we assess how much the chances for a successful attack 
can be improved by replacing random impostor attacks with
skilled impostor attacks.
To this end, we utilize the available databases of 
real fingerprints (DB1, DB2 and DB3) from FVC2000, FVC2002 and FVC2004.
We ignore DB4 of each competition which contains synthetic fingerprint images.

\begin{figure}
\begin{center}
\includegraphics[width=0.9\textwidth]{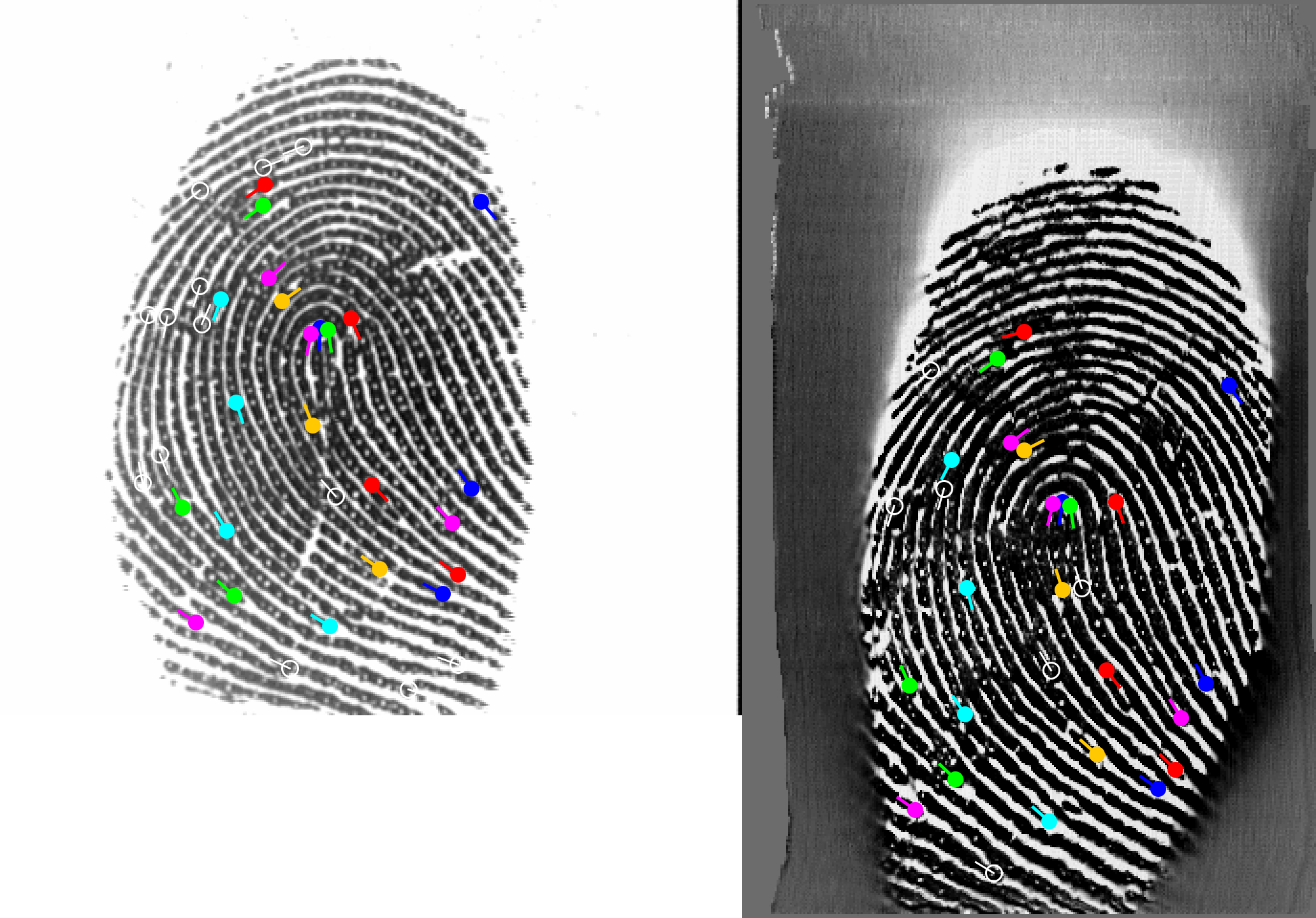} \\
\includegraphics[width=0.9\textwidth]{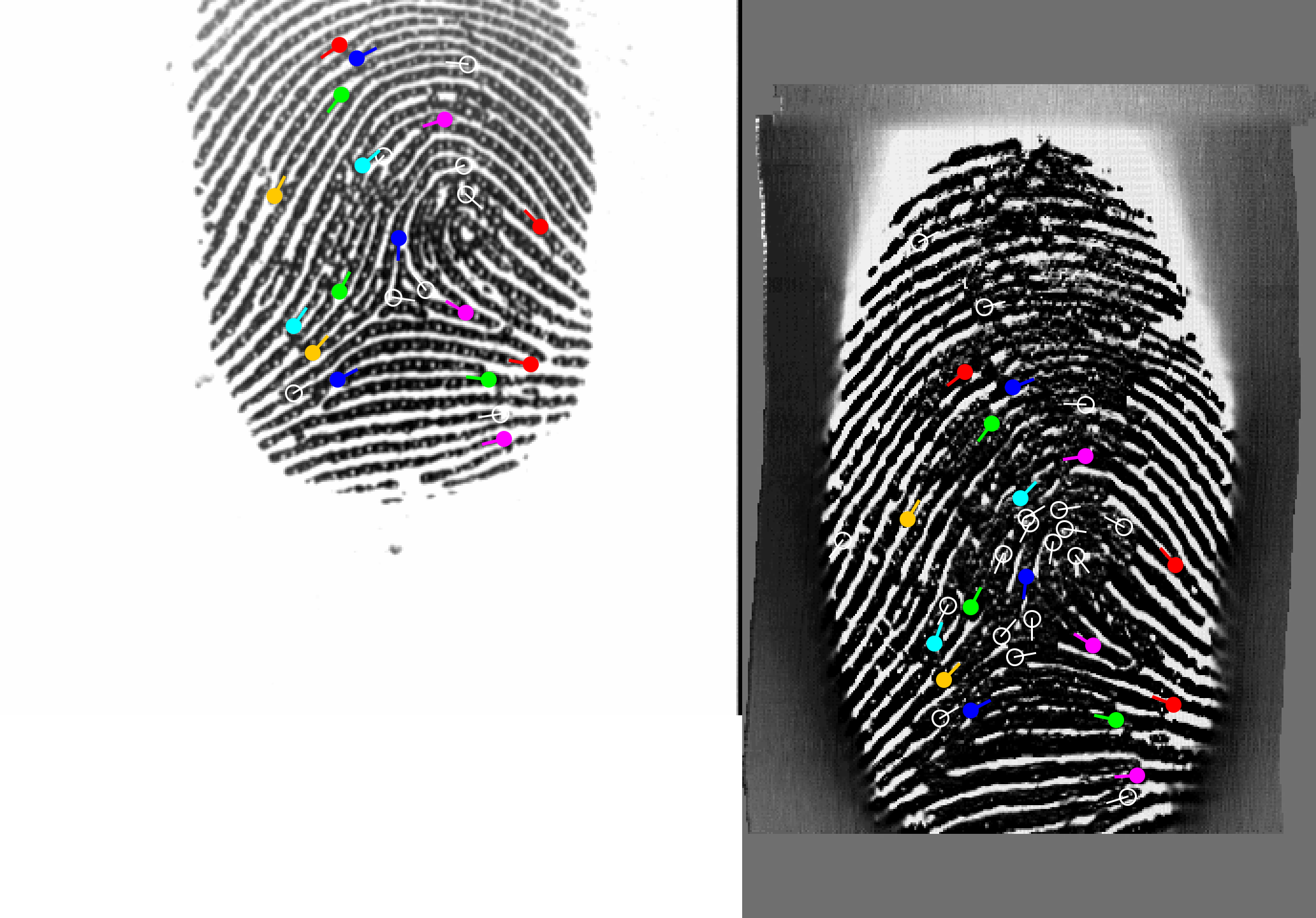}
\end{center}
\caption{A comparison of finger no. 88 impression 5 from FVC2002 DB1 (top left)
and finger no. 43 impression 7 from FVC2004 DB3 (top right) resulted in a score of 267.
Comparing finger no. 81 impression 2 from FVC2002 DB1 (bottom left)
to finger no. 41 impression 8 from FVC2004 DB3 (bottom right) lead to a score of 155.
Paired minutiae are visualized in color, unpaired in white.
}
\label{fig:sameFinger}
\end{figure}

The databases of 
real fingerprints (DB1, DB2 and DB3) from FVC2000, FVC2002 and FVC2004
contain in total 990 fingers. 
Interestingly, the skilled impostor attack revealed 
not all fingers are unique across different FVC databases. 
Using the fingers in FVC2002 DB1 as targets
we discovered that in some cases exactly the same finger appeared also in a second, different database.
Two examples of duplicates are shown in Figure \ref{fig:sameFinger}.
The discovered duplicates have been excluded from the verification test.

Mallory can improve her chances of a successful attack 
on Alice's bank account from $0.1\%$ for random impostor attacks
to $88.875\%$ for skilled impostor attacks.
We have observed that in 711 out of 800 attempts
a score of 48 or greater has been achieved by the skilled impostor.
A score of $s = 120$ has been computed for the example shown in Figure \ref{fig:skilledImpostor}.
The verification performance for this test with skilled impostors is depicted as a red detection error tradeoff curve
in Figure \ref{fig:det}.

Moreover, we performed a second verification test with skilled impostor attacks 
against the same 800 images of FVC2002 DB1.
For the attack, we used a second database which contains more fingers 
than all FVC databases combined and the image quality is also higher.
In this second test, all 800 skilled impostor attacks were successful.

These results demonstrate how important it is to take skilled impostor attacks into account
when determining the operational threshold for a fingerprint verification system.

\section{Conclusion}

We have described a new attack vector on fingerprint verification systems
denoted as skilled impostor attack.
In the existing literature, only random impostor attacks (zero-effort attacks)
and spoof attacks have been considered so far as attacks against verification systems.
In addition to traditional verification protocols with random impostor attacks,
we have proposed a new verification protocol which performs skilled impostor attacks.
This protocol should be used in evaluations of fingerprint verification performance
and for determining the operational threshold of a system.
If such choices are based on the results of traditional verification tests only,
the operating company 
is in the dark about the possibilities and capabilities of potential attackers
and takes the risk to enormously underestimate the chances of a successful attack.
Such a fingerprint verification system might be vulnerable to skilled impostor attacks.
The results of verification tests with
skilled impostor attacks allow 
a more realistic assessment of potential vulnerabilities 
and enable a more informed decision regarding a trade-off between security (low false acceptance rates)
on the one hand, and usability and convenience for the user (low false rejection rates) on the other hand.

In the light of an increasing pervasiveness of fingerprint recognition systems 
and a strong trend towards authenticating financial transaction by fingerprint verification,
honest evaluations of factual system performance and weaknesses are needed.

Topics deserving further research include 
possible measures to increase the security,
e.g. by combining multiple fingers for a verification decision,
or by combining multiple modalities. 
Especially finger vein recognition \cite{KaubaReissigUhl2014}
is a promising candidate, because first public tests indicate 
that this modality has the potential to achieve 
a comparable performance as fingerprint verification
and sensors exist which acquire 
fingerprint and finger vein images simultaneously \cite{RaghavendraRajaSurbiryalaBusch2014}.
Such sensors can be integrated into an ATM,
whereas an integration into smartphones for mobile payment may be more challenging.

Moreover, research on biometric cryptosystems 
should be considered as a potential source for solutions
which aim for a protected storage of fingerprint data.
Main objectives of biometric cryptosystems are irreversibility,
unlinkability and revocability.
A survey on approaches to achieve these goals is given in \cite{RathgebUhl2011}.
Security considerations and vulnerabilities of biometric cryptosystems 
are analyzed and discussed in \cite{Tams2012PhD,TamsMihailescuMunk2015}.

In conclusion, we would like to emphasize that the skilled impostor attack
is independent of the choice of a specific fingerprint database or a specific comparison algorithm.
In fact, the described attack and the proposed protocol for evaluating verification performance 
is even independent from its application to fingerprint recognition.
A future research direction is to transfer the idea of 
skilled impostor attacks to other modalities like e.g. finger vein, speaker, face or iris recognition
and to perform verification tests according to the new protocol with skilled impostors.

We believe that all biometric modalities which can be captured 
at a distance (fingerprint, face, iris, ear \cite{PflugBusch2012}, gait)
or leave traces behind (fingerprint, signatures)
are especially vulnerable to this attack, 
whereas modalities like finger vein or electroencephalogram (EEG) based biometric recognition \cite{CampisiLarocca2014}
are less susceptible to this kind of attack.

\section*{Acknowledgements}

The author would like to thank Hanno Coetzer for his kind permission to use 
the images shown in Figure \ref{fig:signature}.
C. Gottschlich gratefully acknowledges the support of the 
Felix-Bernstein-Institute for Mathematical Statistics in the Biosciences 
and the Niedersachsen Vorab of the Volkswagen Foundation.


\end{document}